\documentclass{article}

\usepackage{microtype}
\usepackage{graphicx}
\usepackage{subcaption}
\usepackage{booktabs} % for professional tables
\usepackage[table,xcdraw]{xcolor}  % 导入 xcolor 包以支持表格颜色
\usepackage{booktabs}       % professional-quality tables
\usepackage{amsfonts}       % blackboard math symbols
\usepackage{nicefrac}       % compact symbols for 1/2, etc.
\usepackage[utf8]{inputenc} % allow utf-8 input
\usepackage[T1]{fontenc}    % use 8-bit T1 fonts
\usepackage{multirow}       % 导入 multirow 包
\usepackage{graphicx}
\usepackage{amsmath} % 包含数学增强功能
\usepackage{wrapfig}  % 用于文字环绕图片
\usepackage{lipsum}   % 用于生成示例文本
\usepackage{arydshln} % 导入支持虚线的宏包
\usepackage{hyperref}
\hypersetup{breaklinks=true}
\usepackage{pifont}
\hypersetup{
    colorlinks=true,   % Enable colored links
    linkcolor=blue,    % Set the color of internal links
    urlcolor=blue,     % Set the color of external links (URLs)
}
\usepackage{algorithm}
\usepackage{algpseudocode}
\usepackage{amsmath}
\usepackage{amssymb}

\usepackage{amsmath,amsthm}
\usepackage{bm}

\usepackage[preprint]{neurips_2026}
\usepackage{amsmath}
\usepackage{amssymb}
\usepackage{mathtools}
\usepackage{amsthm}

\usepackage[capitalize,noabbrev]{cleveref}

\theoremstyle{plain}

\theoremstyle{definition}

\theoremstyle{remark}

\usepackage[disable,textsize=tiny]{todonotes}

\title{AdaTok: Self-Budgeting Image Tokenization with Quality-Preserving Dynamic Tokens}

\author{%
\normalfont
Xiaocheng Lu$^{1}$ \quad Yuxi Chen$^{1}$ \quad Jie Zhang$^{1}$ \quad Jian Liu$^{1}$ \\
Jingcai Guo$^{2}$ \quad Fangqi Zhu$^{1}$ \quad Tao Han$^{1}$ \quad Song Guo$^{1}$ \\
{\small $^{1}$The Hong Kong University of Science and Technology} \\
{\small $^{2}$The Hong Kong Polytechnic University}
}

\begin{document}
\maketitle

\begin{abstract}
Image tokenizers, from 2D grids to recent 1D sequences, typically encode every image with the same fixed number of tokens. Yet visual complexity is highly heterogeneous, so a uniform budget overspends on simple inputs and underserves complex ones. Existing elastic tokenizers expose variable-length reconstructions, but often leave token length as a deployment-time operating point, a search target, or an external prediction rather than an output of the tokenizer itself. In this work, we ask whether a discrete visual tokenizer can \emph{budget itself} in one pass. Our central finding is that actionable elasticity requires a representation--allocation co-design: prefixes must remain decodable across budgets, and the tokenizer must learn which prefix each image needs. We propose \textbf{AdaTok}, a self-budgeting discrete 1D tokenizer. AdaTok combines \textit{Prioritized Representation Learning}, which orders tokens with nested tail masking and resolves budget-dependent semantic shift through Multi-Head LoRA decoder heads, with \textit{Adaptive Token Allocation}, which trains a lightweight deterministic-group GRPO policy over candidate budgets. Dynamic Pareto Weighting balances fidelity and efficiency during policy training without manual trade-off sweeps. On ImageNet-1K, AdaTok-Full reaches rFID \textbf{1.31} at 256 tokens, while AdaTok-Adaptive attains rFID \textbf{1.50} using only $\sim$118 tokens on average, outperforming discrete 1D baselines at comparable budgets. In autoregressive image generation, the shorter adaptive representation yields $\sim$$2.1{\times}$ throughput over a fixed 256-token decode, suggesting that visual token count can be learned as a content-conditioned output rather than set as a fixed hyperparameter.
\end{abstract}

\section{Introduction}
\enlargethispage{2\baselineskip}

\begin{figure}[t]
  \centering
  \begin{subfigure}[b]{0.68\linewidth}
    \centering
    \includegraphics[width=\linewidth]{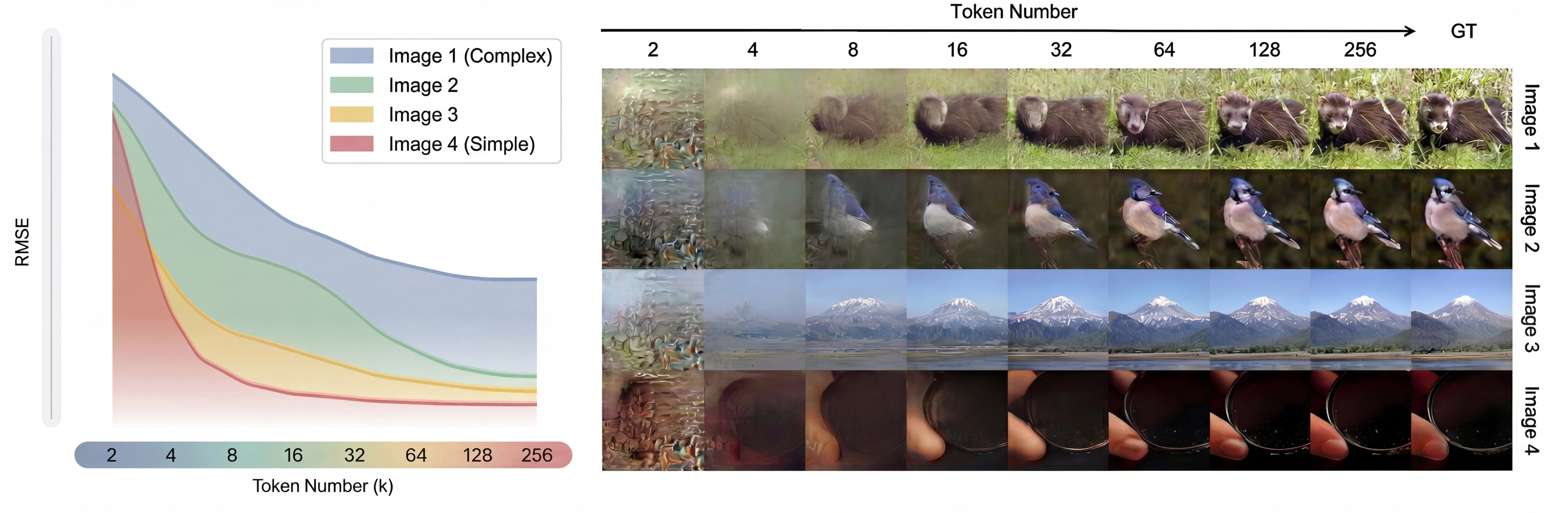}
    \caption{rMSE \emph{vs.} token usage (left) and qualitative reconstructions (right).}
    \label{fig:relation}
  \end{subfigure}
  \hfill
  \begin{subfigure}[b]{0.3\linewidth}
    \centering
    \includegraphics[width=\linewidth]{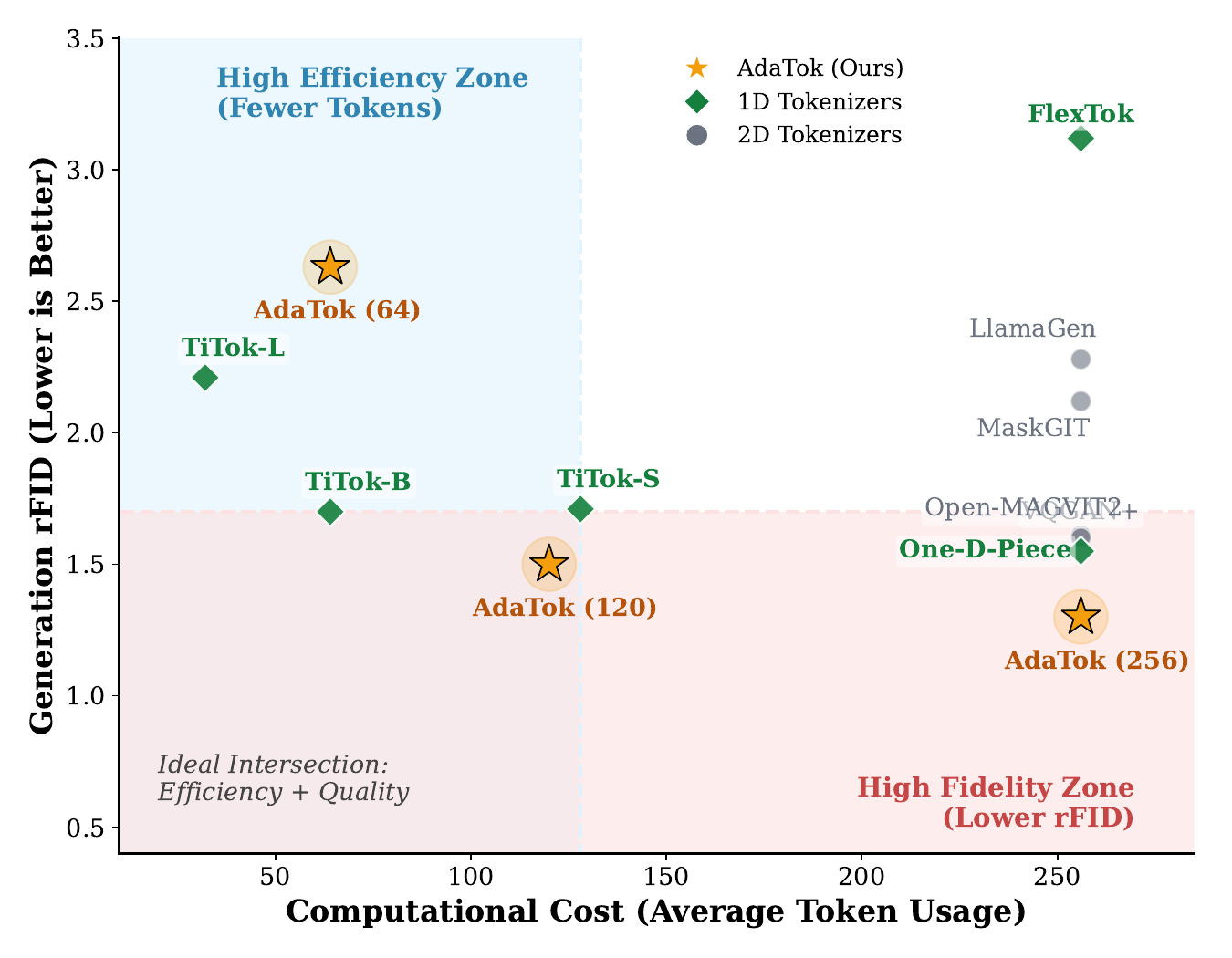}
    \caption{Token--rFID Pareto frontier.}
    \label{fig:quadrant}
  \end{subfigure}
  \caption{\textbf{Motivation.} (a) Reconstruction quality scales non-uniformly with token length across heterogeneous samples, motivating per-instance adaptive allocation rather than a one-size-fits-all budget. (b) AdaTok lies in a favorable quality--efficiency region compared with fixed-budget baselines.}
  \label{fig:teaser}
  \vspace{-12pt}
\end{figure}

Modern image tokenizers, from 2D-grid models such as VQ-VAE~\citep{van2017neural} and VQGAN~\citep{esser2021taming} to recent 1D variants such as TiTok~\citep{yu2025image}, FlexTok~\citep{bachmann2025flextok}, and One-D-Piece~\citep{miwa2025one}, usually make one hidden commitment: every image receives the same token budget. This commitment is convenient for batching and downstream sequence modeling, but it is poorly matched to visual data. A centered logo, a smooth background, and a dense street scene do not require the same description length. Fig.~\ref{fig:teaser}~(a) makes this mismatch concrete: reconstruction quality scales non-uniformly with token length across samples, and the same token count produces markedly different rMSE across image classes. The cost is structural. In autoregressive image generation, latency grows with sequence length; in vision--language models, visual tokens compete with text for context. A fixed-length tokenizer therefore pays a near worst-case price on every input, even when many images admit a shorter discrete description.

Recent work has begun to loosen this rigidity. 1D tokenizers decouple latent tokens from spatial grids~\citep{yu2025image}, and elastic representations reconstruct images from different prefix lengths~\citep{bachmann2025flextok, miwa2025one}. Other adaptive tokenizers use recurrent or search-based allocation~\citep{yan2024elastictok, duggal2024adaptive}, ordered latent importance~\citep{wen2025principal}, or external complexity predictors~\citep{shen2025cat}. These methods show that variable-length visual representation is possible, but they leave a gap between \emph{flexible reconstruction} and \emph{autonomous tokenization}. First, variable-length decoding does not specify which length an input should use; the budget is still chosen by a fixed operating point, test-time search, target-quality conditioning, or an external model. Second, a nested prefix is not automatically reusable across budgets. The same leading tokens must act as a standalone structural summary at low budgets and as a coarse scaffold at high budgets, creating a budget-dependent semantic shift for a shared decoder. The missing object is therefore an actionable elastic tokenizer: a discrete prefix stream that remains decodable at each budget, paired with an internal one-pass policy that decides how many tokens the image needs.

We propose \textbf{AdaTok}, a self-budgeting discrete 1D tokenizer that turns token count into a learned, input-conditioned output. Our central finding is that actionable elasticity is a representation--allocation co-design problem. As shown in Fig.~\ref{fig:framework}, AdaTok has two coupled components.
\textbf{Prioritized Representation Learning (PRL)} builds an ordered prefix stream with Nested Tail Masking and equips the decoder with budget-specific Multi-Head LoRA heads, so each prefix is decoded under the modulation appropriate to its semantic role.
\textbf{Adaptive Token Allocation (ATA)} then freezes the tokenizer and trains a lightweight policy with Group Relative Policy Optimization (GRPO). For each image, the candidate budget set itself forms a deterministic comparison group, giving an instance-relative advantage without sampling variance or a value network. Dynamic Pareto Weighting (DPW) further balances the fidelity and length rewards by their gradient influence, avoiding a manual sweep over scalarization weights.

We validate the full chain across reconstruction, ablation, and downstream generation. On ImageNet-1K, AdaTok-Full reaches rFID \textbf{1.31} at 256 tokens, while AdaTok-Adaptive reaches rFID \textbf{1.50} at only $\sim$118 average tokens, outperforming the discrete 1D baselines we compare against at comparable budgets. Ablations show that Nested Tail Masking alone hurts reconstruction, but MH-LoRA resolves the semantic shift and makes ordered prefixes usable. On the allocation side, deterministic-group GRPO with DPW improves both quality and average length over heuristic allocation, REINFORCE, PPO, and GRPO without dynamic weighting. In autoregressive generation, the 118-token operating point yields $\sim$$2.1{\times}$ throughput over a fixed 256-token decode, and matched-compute analysis shows that shorter prefixes can preserve or even improve generation quality in the appropriate regime.
Our contributions are summarized as follows:
\begin{itemize}
    \item \textbf{Actionable elasticity for discrete tokenizers.} We turn variable-length reconstruction into self-budgeted tokenization: the tokenizer learns an internal per-image budget decision and keeps the selected discrete prefix decodable via budget-specialized MH-LoRA.
    \item \textbf{Deterministic-group GRPO for allocation.} We adapt GRPO by using the discrete candidate budget set as the policy group itself, giving an exact instance-relative advantage with no sampling variance, no critic, and no external predictor; Dynamic Pareto Weighting auto-balances fidelity against length.
    \item \textbf{Quality-preserving token reduction.} AdaTok reaches rFID \textbf{1.50} at $\sim$118 average tokens, a favorable reconstruction point among the discrete 1D tokenizers we compare against, and delivers $\sim$$2.1{\times}$ AR-generation throughput over a fixed 256-token baseline.
\end{itemize}

\section{Related Work}
\enlargethispage{2\baselineskip}

Compressing high-resolution images into compact latent spaces is foundational to modern visual generation. Continuous tokenization via VAEs~\citep{kingma2013auto} underpins latent diffusion models~\citep{rombach2022high}, which perform denoising in a latent manifold. In parallel, discrete tokenization, pioneered by VQ-VAE~\citep{van2017neural} and refined by VQGAN~\citep{esser2021taming}, quantizes images into a 2D grid of discrete codes. While dominant, the 2D-grid layout imposes a rigid structural prior that wastes capacity on simple regions and underserves complex ones.

To break this rigidity, recent work has shifted toward structure-free 1D sequences. TiTok~\citep{yu2025image} showed that an image can be represented as a compact 1D sequence; FlexTok~\citep{bachmann2025flextok} and One-D-Piece~\citep{miwa2025one} extended this to variable-length encoding. Beyond elastic representations, several works study how to allocate the budget itself: Semanticist~\citep{wen2025principal} orders latent tokens by information importance, CAT~\citep{shen2025cat} delegates the choice to an external LLM-based complexity predictor, and ElasticTok~\citep{yan2024elastictok} performs test-time search. Concurrent work KARL~\citep{duggal2025karl} also predicts token count in a single forward pass, via Upside-Down RL conditioned on a target reconstruction quality. AdaTok targets a different cell in this design space: discrete 1D codes, elastic prefixes, internal one-pass allocation, no search or external predictor, no target-quality conditioning, and a budget-specialized decoder that makes the selected prefix usable.

The work also connects to adaptive computation in Vision Transformers such as DynamicViT~\citep{rao2021dynamicvit} and A-ViT~\citep{yin2022vit}, which prune redundant tokens at inference. Where those methods accelerate specific backbone layers, AdaTok aims at a reusable adaptive representation: by framing token selection as a multi-objective decision optimized with Group Relative Policy Optimization (GRPO), AdaTok provides a single-pass alternative to search-based allocation and aligns naturally with the variable-length structure of language models~\citep{sun2024autoregressive, tian2024visual, yang2022visual}.

\section{Preliminaries and Motivation}
\label{sec:method_theory}

\textbf{Setup.} Let $\mathbf{x} \in \mathbb{R}^{H \times W \times C}$ denote an input image. A canonical discrete tokenizer maps $\mathbf{x}$ to a latent sequence $\mathbf{z} = [z_1, \dots, z_N] \in \mathcal{V}^N$ via an encoder $E$ and quantizer $Q$, then reconstructs $\hat{\mathbf{x}}$ via a decoder $G$. \textbf{Adaptive Tokenization} replaces the fixed length $N$ with an instance-dependent length $k \leq N$ via a policy $\pi_\theta(k|\mathbf{x})$, yielding a truncated sequence $\mathbf{z}_{1:k}$.

\textbf{Why fixed budgets are wasteful.} Two classical results from information theory motivate adaptivity. First, in the lossless regime, Shannon's source coding theorem implies that the optimal code length for a sample $x$ scales as $K^\star(x) \propto -\log p(x)$, so a fixed budget $K(x)\equiv N$ overspends on simple samples and underserves complex ones; the bound is conceptually analogous to (rather than directly equivalent to) the lossy rate--distortion case relevant to image tokenization. Second, in the lossy regime, the rate--distortion bound requires a worst-case fixed-budget tokenizer to spend capacity according to the most complex instances, even when the average image needs fewer tokens. Together these results call for per-instance allocation rather than a uniform worst-case budget that pays the same cost on every input.

\textbf{Why a single fixed weighting is hard to optimize.} Adaptive tokenization has two competing objectives: distortion $\mathcal{D}(\mathbf{x},\pi_\theta)$ and length $K(\mathbf{x};\theta)$. A single global scalar loss $w_1\mathcal{D}+w_2 K$ can describe an aggregate first-order trade-off, but the right $\mathbf{w}$ is hard to find empirically. In practice, the per-instance marginal value of additional tokens varies sharply, the length gradient is dominated by majority easy images, and tuning $\mathbf{w}$ to a target operating point requires extensive sweeping. AdaTok sidesteps this with a per-instance baseline that produces a stable scale-invariant relative-reward signal (GRPO + DPW, Sec.~\ref{sec:ata}).

\begin{figure*}[t]
  \centering
  \includegraphics[width=0.9\textwidth]{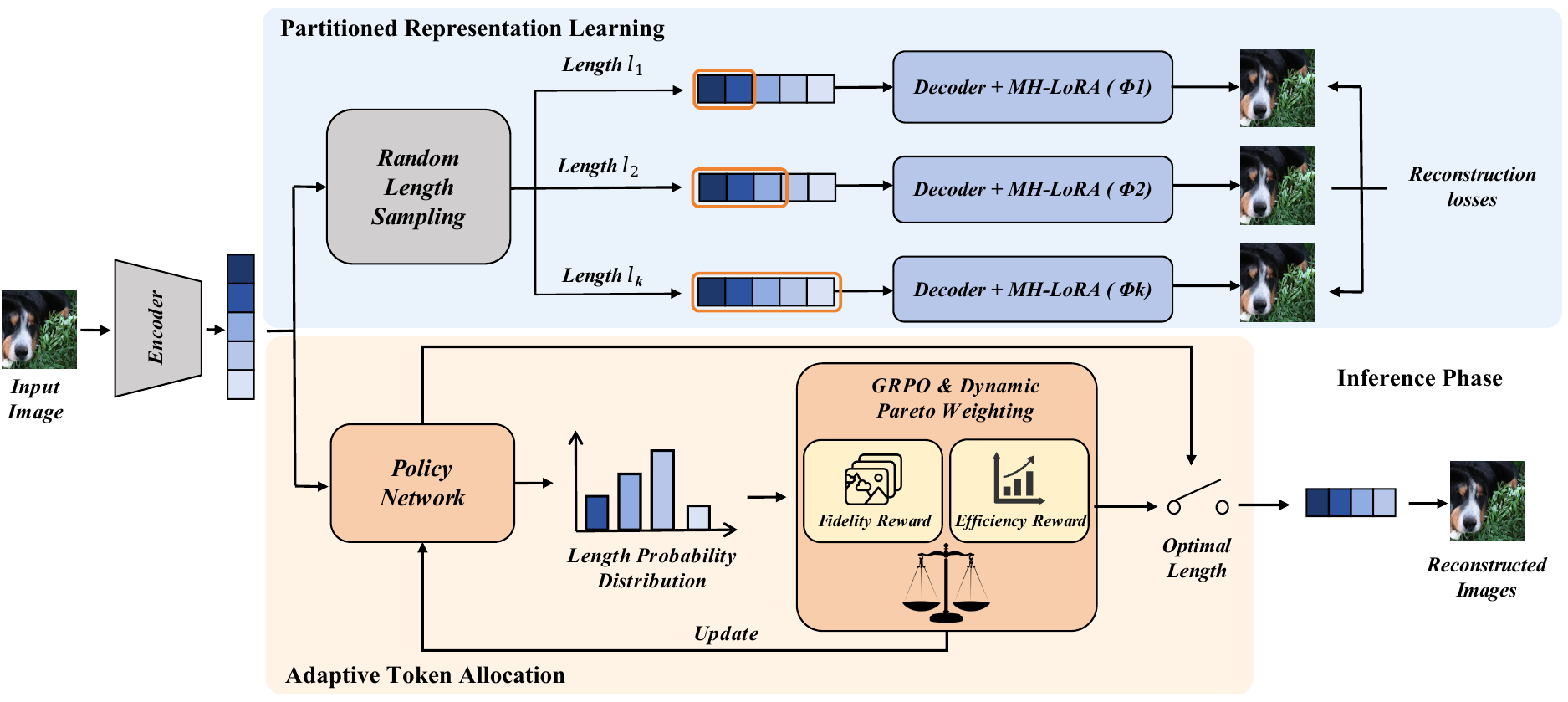}
   \caption{\textbf{Framework of AdaTok.} AdaTok integrates Prioritized Representation Learning (PRL) to learn hierarchical 1D tokens and an Adaptive Token Allocation (ATA) policy to autonomously select a content-adaptive token budget for each image.}
   \label{fig:framework}
   \vspace{-10pt}
\end{figure*}

\section{Method}
Section~\ref{sec:method_theory} identifies two limits of static tokenization: fixed budgets create systematic redundancy, and the fidelity--efficiency trade-off varies across inputs. An adaptive tokenizer therefore needs two capabilities: a latent space that preserves an information hierarchy across budgets, and a policy that chooses an input-specific operating point.

Following these principles, \textbf{AdaTok} extends 1D image tokenization into a self-budgeting tokenizer: length is predicted by the tokenizer itself, not chosen after training by a user-defined operating point. As shown in Fig.~\ref{fig:framework}, it has two coupled modules.
\textit{Prioritized Representation Learning (PRL)} builds the length-aware representational foundation: nested masking enforces an ordered information hierarchy, and budget-specific decoder modulation guarantees that any prefix of the 1D latent sequence remains a high-quality representation of the input.
\textit{Adaptive Token Allocation (ATA)} is the decision engine: a lightweight policy network trained with Group Relative Policy Optimization (GRPO) picks a content-adaptive token budget in one forward pass, without test-time search or external guidance.

\subsection{Prioritized Representation Learning}
\label{sec:prl}

\textbf{Information Ordering via Nested Tail Masking.}
Standard discrete tokenizers treat all latent positions as equally important, which produces semantic redundancy. Following 1D tokenizers~\citep{yu2025image}, we represent an image as a sequence of $N$ latent tokens. To make prefixes actionable rather than merely inspectable, we apply \textit{Nested Tail Masking (NTM)} during training~\citep{rippel2014learning, miwa2025one}: given a latent sequence $\mathbf{z} \in \mathbb{R}^{N \times d}$, we sample a budget $l \in \mathcal{L}$ and mask the trailing tokens $[l+1, N]$. This forces the encoder to pack the most invariant, information-dense semantics into the prefix and to use the trailing tokens for reconstruction refinement, yielding an \textit{implicit bit-stream} in which early tokens carry the semantic foundation and later tokens add fidelity.

\textbf{Decoupling Semantic Shift via Multi-Head LoRA.}
NTM yields a flexible latent sequence, but it also creates a \textit{semantic shift} problem. The same prefix tokens must summarize the image when the budget is small and act as a coarse scaffold when the budget is large. This is not merely a capacity issue: the conditional meaning of a prefix changes with the chosen budget. A static decoder, as used by existing elastic tokenizers~\citep{miwa2025one, bachmann2025flextok}, has no way to separate these roles and is forced into a sub-optimal compromise.

We address this with \textbf{Multi-Head LoRA (MH-LoRA)} inside the decoder. The budget space is discretized into $K{=}8$ buckets $\mathcal{L} = \{32, 64, 96, 128, 160, 192, 224, 256\}$, and each bucket gets a dedicated rank-$r$ LoRA head $\phi_k$ (we use $r{=}16$, adding $\sim$1.05M parameters or $\sim$1.1\% of the model). During training, NTM samples a budget $l_k \in \mathcal{L}$ and the reconstruction becomes
\begin{equation}
    \hat{\mathbf{x}} = \text{Decoder}\!\left(Q(\mathbf{z}_{1:l_k});\, \Theta + \phi_k\right),
\end{equation}
where $\Theta$ is the shared base decoder and $\phi_k$ is the budget-specific modulation. \textbf{At inference}, the ATA policy (Sec.~\ref{sec:ata}) returns a categorical distribution over $\mathcal{L}$; we take the argmax $k^\star$ and decode with $\phi_{k^\star}$ on the truncated sequence $\mathbf{z}_{1:l_{k^\star}}$. By specializing only lightweight residual heads, MH-LoRA makes each prefix decodable at its own semantic scale while preserving a shared codebook and base decoder.

\textbf{Optimization Objective.}
The PRL module is trained end-to-end to minimize the reconstruction loss across all possible budgets in $\mathcal{L}$:
\begin{equation}
\begin{aligned}
    \mathcal{L}_{\text{PRL}} = \mathbb{E}_{l \in \mathcal{L}} \big[ & \mathcal{L}_{\text{recon}}(\mathbf{x}, \hat{\mathbf{x}}_l) + \lambda_p \mathcal{L}_{P}(\mathbf{x}, \hat{\mathbf{x}}_l) \\
    & + \lambda_g \mathcal{L}_{G}(\hat{\mathbf{x}}_l) + \mathcal{L}_{\text{commit}} \big]
\end{aligned}
\end{equation}
where $\mathcal{L}_{\text{recon}}$, $\mathcal{L}_P$, and $\mathcal{L}_G$ are the pixel-level, perceptual, and adversarial losses, respectively. Optimizing across all budgets jointly forces the shared codebook to encode multi-granularity visual features rather than overfitting to a single length.

\subsection{Adaptive Token Allocation (ATA)}
\label{sec:ata}
On top of the PRL foundation, ATA learns a lightweight policy network $\pi_{\theta}$ that picks a content-adaptive token length $l$ for each input image.

\textbf{Group Relative Policy Optimization.}
We frame token selection as a discrete decision problem. Because the fidelity--length trade-off varies sharply across images, standard policy-gradient methods exhibit high variance: easy images receive high rewards regardless of allocation, obscuring the marginal value of additional tokens. We therefore adapt \textit{Group Relative Policy Optimization} (GRPO)~\citep{shao2024deepseekmath} to the elastic-tokenization setting, with a key twist: the \emph{group itself is the discrete candidate budget set $\mathcal{L}$}, evaluated deterministically on the same image. The policy therefore compares ``how many tokens does this image need?'' against all available budgets and obtains an exact relative advantage per budget without sampling variance. The choice has two further benefits. First, \textit{content-dependent baseline calibration}: normalizing rewards by the group mean removes per-image difficulty bias and lets the policy focus on the marginal gain from additional tokens. Second, GRPO realizes instance-dependent adaptation without an auxiliary value network, keeping inference lightweight.

\textbf{Rate--Distortion Reward Formulation}.
For each $l \in \mathcal{L}$ in the deterministic group, the rewards are the fidelity term $r_{\text{mse}}(l) = -\,\text{MSE}(\mathbf{x}, \hat{\mathbf{x}}_l)$ and the efficiency term $r_{\text{len}}(l) = -\,l / l_{\max}$. We do not apply an explicit reward scale: under group-relative normalization any positive multiplier cancels in the standardized advantage (up to the stability constant $\epsilon$).
GRPO computes the relative advantage $A$ by standardizing these rewards over the group $\mathcal{L}$ (deterministic, no sampling):
\begin{equation}
    A_{r}(l, \mathbf{x}) = \frac{r(l, \mathbf{x}) - \mu_{\mathcal{L}}(r)}{\sigma_{\mathcal{L}}(r) + \epsilon}.
\end{equation}
Here $\mu_{\mathcal{L}}, \sigma_{\mathcal{L}}$ are the per-image mean and standard deviation across the candidate set. We interpret the resulting policy as a \emph{content-conditioned} allocation rule rather than a strict per-instance R--D optimum, since standardization removes absolute distortion magnitudes.

% \textbf{Dynamic Pareto Weighting}.
% While GRPO handles per-instance adaptation, balancing the global convergence between fidelity and efficiency remains a challenge. We employ \textit{Dynamic Pareto Weighting} to stabilize the multi-objective optimization. The ATA module minimizes the following objective:
% \begin{equation}
% \begin{aligned}
%     \mathcal{L}_{\text{ATA}} = &-\sum_{l \in \mathcal{L}} \pi_{\theta}(l|z) \left( w_{\text{mse}} A_{\text{mse}}(l) + w_{\text{len}} A_{\text{len}}(l) \right) \\
%     &- \beta H(\pi_{\theta})
% \end{aligned}
% \end{equation}
% where $H(\pi_{\theta})$ is entropy regularization to encourage exploration. To avoid manual tuning, the weights $\mathbf{w} = [w_{\text{mse}}, w_{\text{len}}]$ are dynamically adjusted based on the relative gradient norms to keep both objectives on a comparable scale during training:
% \begin{equation}
%     w_i = \frac{\exp(\eta_i)}{\sum_j \exp(\eta_j)}, \quad \eta_i \leftarrow \eta_i + \gamma \cdot \|\nabla_{\theta} \mathcal{L}_i\|
% \end{equation}
\textbf{Dynamic Pareto Weighting}.
While GRPO handles per-instance adaptation, balancing the global convergence between fidelity and efficiency remains a challenge. We employ \textit{Dynamic Pareto Weighting} (DPW) to stabilize the multi-objective optimization. The ATA module minimizes the following objective:
\begin{equation}
\begin{aligned}
    \mathcal{L}_{\text{ATA}} = &-\sum_{l \in \mathcal{L}} \pi_{\theta}(l|z) \left( w_{\text{mse}} A_{\text{mse}}(l) + w_{\text{len}} A_{\text{len}}(l) \right) \\
    &- \beta H(\pi_{\theta})
\end{aligned}
\end{equation}
where $H(\pi_{\theta})$ is entropy regularization to encourage exploration, and we denote the individual objectives as $\mathcal{L}_i = - \sum_{l \in \mathcal{L}} \pi_\theta(l|z) A_i(l)$ for $i \in \{\text{mse}, \text{len}\}$. To avoid manual tuning and encourage synergistic learning, the weights $\mathbf{w} = [w_{\text{mse}}, w_{\text{len}}]$ are dynamically adjusted based on the gradient influence of each objective:
\begin{equation}
    I_i^{(t)} = \Big\langle \nabla_{\theta} \mathcal{L}_i^{(t)}, \sum_k \nabla_{\theta} \mathcal{L}_k^{(t)} \Big\rangle, \quad w_i^{(t)} = \frac{w_i^{(t-1)} \exp\left( \frac{\eta}{\mu} I_i^{(t)} \right)}{\sum_j w_j^{(t-1)} \exp\left( \frac{\eta}{\mu} I_j^{(t)} \right)}
\end{equation}
Here $\eta$ is the ATA learning rate (distinct from the reward scale $\alpha$) and $\mu$ is a regularization factor; we use $\mu{=}10$, and sweeps over $\{1,5,10,50\}$ changed final rFID by less than 0.05. The update emphasizes the objective with the larger marginal contribution to joint progress~\citep{lu2025learning}, improving the empirical rate--distortion trade-off while preserving single-pass budget selection (Tab.~\ref{tab:ablation}).

\textbf{Training Pipeline.}
We adopt a two-stage scheme rather than joint end-to-end optimization: Stage 1 trains the tokenizer (encoder, decoder, MH-LoRA heads, codebook) to convergence; Stage 2 freezes it and trains the ATA policy on top. Joint training is fragile for three reasons. (i)~\textit{Reward non-stationarity}: GRPO's reward is reconstruction quality, but during early tokenizer training the short-length reconstructions are noisy and the policy collapses to selecting the maximum budget. (ii)~\textit{Codebook fragility}: VQ codebook learning is notoriously brittle, and injecting RL gradients before the codebook stabilizes risks collapse; our two-stage scheme attains $>$98.8\% codebook utilization. (iii)~\textit{Objective conflict}: high reconstruction across all lengths and minimum-token allocation are inherently competing goals, which makes joint convergence hard to control. Decoupling lets the tokenizer first build a clean latent space, after which the policy learns allocation on a stationary reward in only 10K steps ($\sim$3\% of total training cost).

\begin{table*}[t]
\centering
\caption{\textbf{Reconstruction and Generation Across Tokenizers.} We report AdaTok at two operating points: \textbf{AdaTok-Full} uses the full 256-token budget and demonstrates the capacity ceiling of our PRL-trained tokenizer; \textbf{AdaTok-Adaptive} uses the per-image budget chosen by the ATA policy (118 tokens on average) and demonstrates the rate--distortion advantage of adaptive allocation. Within each row, all metrics are measured at the same operating point. Bold highlights AdaTok results among the discrete 1D rows. Generation numbers use the generators reported by each method and should be interpreted as compatibility indicators rather than a controlled generator comparison.}
\resizebox{\linewidth}{!}{%
\begin{tabular}{lcccccc|cc}
\hline
\textbf{Tokenizer}          & \textbf{\#Params}             & \textbf{\#Codebook} & \textbf{\#Tokens} & \textbf{Elastic} & \textbf{rFID}$\downarrow$ & \textbf{PSNR}$\uparrow$ &  \textbf{Generator} & \textbf{gFID}$\downarrow$ \\ \hline
\multicolumn{9}{c}{\textit{fixed-budget baselines}}                                                                                                                              \\ \hdashline
VQGAN    &  85M     & 1024                   & 256             & \ding{55}             & 7.94 & 19.4 & LDM-8    & 15.78     \\
MaskGIT        & 227M      & 1024                   & 256             & \ding{55}                  & 2.12 & - & MaskGIT   & 6.18        \\
LlamaGen & 72M  & 16384                  & 256            & \ding{55}                    & 2.28 & 20.79 & LlamaGen  & 3.06 \\
ImageFolder  &  362M  &  4096  &  286     & \ding{55}   &  0.80 & - &VAR-d16   &   2.60  \\
VAR &  310M   & 4096   & 680   & \ding{55}  & 0.90 & - & VAR-d16  &   3.30  \\
TiTok-S  & 83M      & 4096                   & 128          & \ding{55}                  & 1.71  & 17.80 & MaskGIT     & 1.97       \\
TiTok-B   & 204M      & 4096                   & 64            & \ding{55}                   & 1.70    & 17.13   & MaskGIT   & 2.48      \\
TiTok-L   & 641M     & 4096                   & 32            & \ding{55}                  & 2.21  & 15.96  & MaskGIT   & 2.77     \\
\hline
\multicolumn{9}{c}{\textit{continuous-token adaptive baselines (contextual; not directly token-count comparable to discrete 1D)}}                                                       \\ \hdashline
CAT   & 187M  & - & - & \ding{51}  & 0.46 & - & - & -\\
Semanticist  & -  &-  & 256   & \ding{51}  & 0.72 & - & LlamaGen & 2.57 \\
\hdashline
\multicolumn{9}{c}{\textit{discrete-token elastic baselines (directly comparable)}}                                                       \\ \hdashline
FlexTok & 341M  & 64000                   & 32         & \ding{51}                     & 4.20   & -  & FlexTok  & 3.83      \\
ALIT & -    &  4096   &    256 & \ding{51}   & 8.25   & -  & - & -  \\
One-D-Piece   & 83M     & 4096                   & 256           & \ding{51}                   & 1.55  & 18.28 & MaskGIT   & 2.67       \\
\hline
\rowcolor{gray!15}
\textbf{AdaTok-Full}     & 97M & 4096 & 256                  & \ding{51} & \textbf{1.31} & \textbf{18.42} & MaskGIT & \textbf{2.28} \\
\rowcolor{gray!30}
\textbf{AdaTok-Adaptive} & 97M & 4096 & \textbf{118 avg.} & \ding{51} & \textbf{1.50} & 17.78             & --      & -- \\
\end{tabular}%
}
\vspace{-10pt}
\label{tab:rec}
\end{table*}

\section{Experiments}
\label{sec:experiment}

\textbf{Implementation Details.}
\textit{Data and evaluation.} All experiments use ImageNet-1K at $256{\times}256$. Reconstruction metrics (rFID, PSNR, SSIM, LPIPS) are computed on the standard 50K validation split; gFID follows the ADM protocol~\citep{dhariwal2021diffusion} on 50K class-conditional samples.
\textit{Architecture.} Encoder and decoder follow the TiTok-S backbone (8 layers, $d_{\text{model}}{=}512$, 8 attention heads, $16{\times}16$ patches); the codebook has 4096 entries at latent dimension 12; the ATA policy is a 3-layer GeLU MLP on globally pooled encoder features.
\textit{Training.} \emph{Stage 1 (PRL)} trains the tokenizer end-to-end for 200 epochs with AdamW (lr 1e-4, weight decay 0.05, batch size 256) and a linear warm-up; nested tail masking samples a budget per step uniformly over the 8 buckets. \emph{Stage 2 (ATA)} freezes the tokenizer and trains the policy with GRPO for 10K steps ($\sim$1\,h, $\sim$3\% of total cost) at group size 8. Decoupling matters: training the policy on top of an unconverged tokenizer yields non-stationary rewards and the policy collapses to the maximum budget. Hardware: 8$\times$NVIDIA H100.

\subsection{Overall Performance}
We benchmark AdaTok against fixed-budget tokenizers (VQGAN, MaskGIT~\citep{Maskgit}, LlamaGen~\citep{sun2024autoregressive}, the TiTok series) and flexible or adaptive tokenizers (One-D-Piece, FlexTok, Semanticist~\citep{wen2025principal}, LLM-guided CAT~\citep{shen2025cat}, search-based ElasticTok~\citep{yan2024elastictok}, multi-scale VAR~\citep{tian2024visual}, recurrent ALIT~\citep{duggal2024adaptive}), assessing reconstruction fidelity, generation quality, and allocation efficiency.

\textbf{Reconstruction and Generation.}
\textit{Reconstruction (primary metric).} Tab.~\ref{tab:rec} reports AdaTok at two operating points and isolates two distinct claims. First, at the matched 256-token budget, \textbf{AdaTok-Full} reaches rFID \textbf{1.31}, the lowest among the discrete 1D baselines we compare against (vs.\ TiTok-S 1.71, TiTok-B 1.70, TiTok-L 2.21, One-D-Piece 1.55), while using only 15\% of TiTok-L's parameter count (97M vs.\ 641M). Second, when the ATA policy chooses the budget per image, \textbf{AdaTok-Adaptive} attains rFID \textbf{1.50} at an average of \textbf{118} tokens --- still better than every fixed-length discrete 1D baseline at this token cost or higher, and roughly $2.2{\times}$ shorter than One-D-Piece's 256 tokens at a similar quality level. The two rows together demonstrate that PRL provides a strong reconstruction ceiling and ATA delivers a favorable rate--distortion operating point.

\textit{Generation (compatibility check).} Generation is not our optimization target; we report gFID only to confirm that AdaTok tokens compose with standard generators. Because MaskGIT requires a fixed sequence length, gFID is reported at AdaTok-Full's 256-token operating point. Under this protocol AdaTok reaches gFID \textbf{2.28}, a favorable result among adaptive-token baselines that report compatible generator evaluations (Semanticist 2.57, One-D-Piece 2.67, FlexTok 3.83). Fixed-length TiTok-S attains a stronger gFID (1.97), but is locked to 128 tokens at both training and inference; AdaTok instead exposes a continuous rate--quality control without retraining. Closing the residual gap to specialist generators would require joint generator--tokenizer training and a larger compute envelope, which we leave to future work.

\begin{figure}[t]
\centering
\begin{minipage}[t]{0.42\linewidth}
  \centering
  \includegraphics[width=\linewidth]{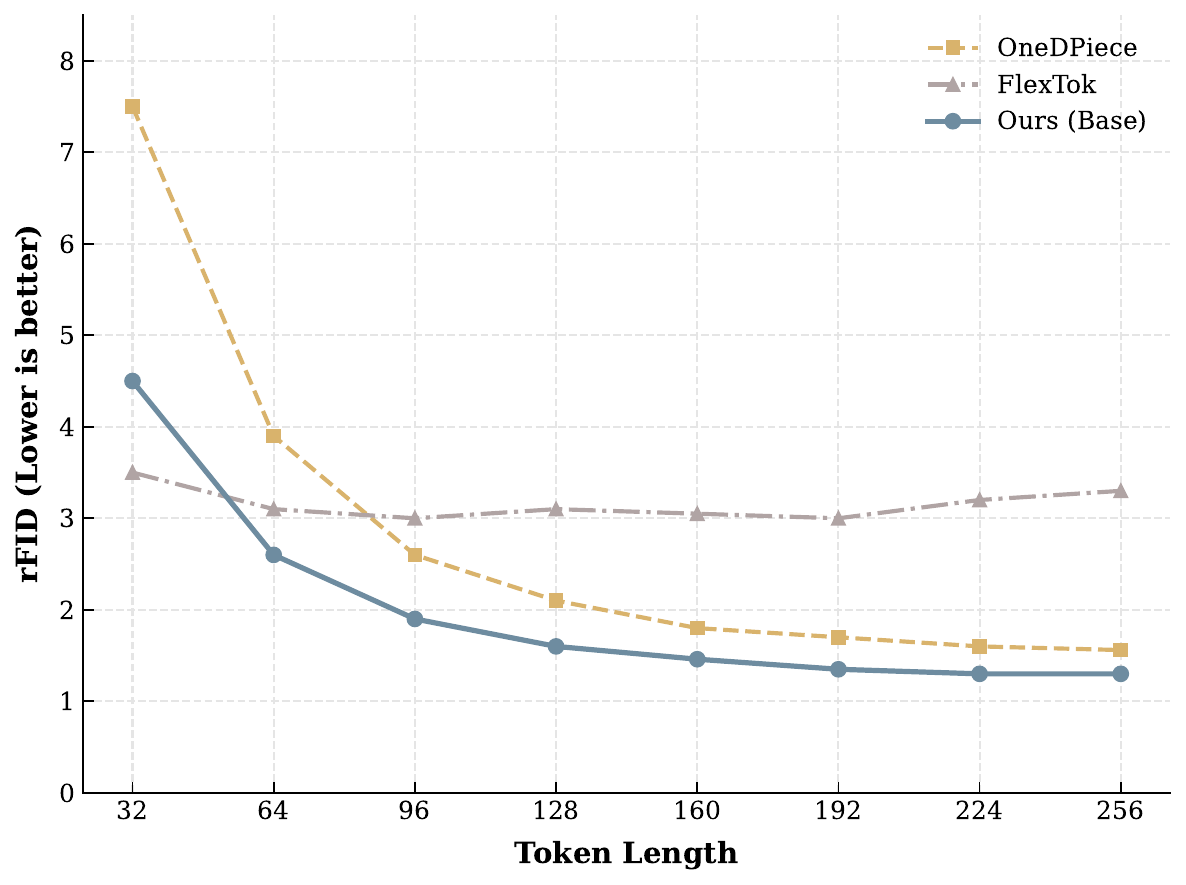}
  \caption{rFID across token budgets for AdaTok versus similarly sized discrete elastic tokenizers (TiTok-S, One-D-Piece, FlexTok-d12-d12). AdaTok stays close to its frontier and remains stable across the full 32--256 token range.}
  \label{fig:elastic}
\end{minipage}\hspace{0.04\linewidth}
\begin{minipage}[t]{0.42\linewidth}
  \centering
  \includegraphics[width=\linewidth]{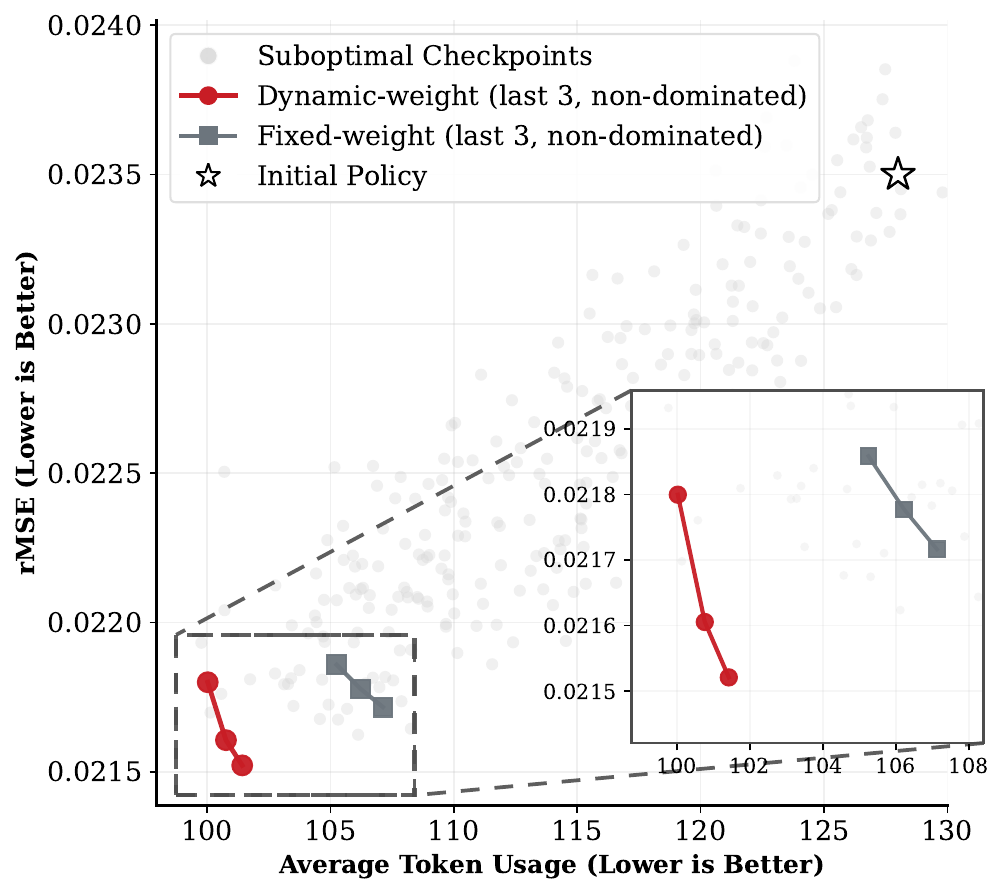}
  \caption{ATA training trajectory in the (avg.\ tokens, rMSE) plane; lower-left is better. GRPO with dynamic Pareto weighting (red) reaches a better fidelity--budget trade-off than fixed-weight scalarization (gray) from the same initial policy.}
  \label{fig:trajectory}
\end{minipage}
\vspace{-10pt}
\end{figure}

\textbf{Elasticity and Adaptivity.}
AdaTok offers two complementary advantages. First, it outperforms existing \emph{discrete} elastic tokenizers (FlexTok, One-D-Piece, ALIT) on both reconstruction and generation. Second, it supports reconstruction at any token count in the 32--256 range without retraining, providing a continuous rate--distortion control knob. Fig.~\ref{fig:elastic} compares AdaTok against similarly sized elastic competitors (TiTok-S, One-D-Piece, FlexTok-d12-d12) across the full budget range: AdaTok leads in the 96--256 region and stays stable elsewhere. FlexTok is competitive in the 32--64 range, but its training objective targets perceptual rather than pixel-level fidelity, making rFID comparisons at very low bitrates nuanced. Continuous adaptive tokenizers (Semanticist 0.72, CAT 0.46) report lower rFID than AdaTok-Adaptive (1.50), but their tokens are float vectors that carry roughly an order of magnitude more bits than a single discrete code from our 4096-entry vocabulary. A fair comparison is therefore bits per pixel rather than token count. We focus on the discrete regime because discrete codes plug directly into autoregressive next-token generators (LlamaGen, MaskGIT) and discrete-vocabulary VLM pipelines without additional quantization loss. Consistent with this focus, AdaTok also leads the adaptive baselines on the directly comparable downstream metric, reaching gFID 2.28 versus 2.57 for Semanticist (Tab.~\ref{tab:rec}). We therefore view AdaTok as a competitive \emph{discrete} adaptive tokenizer with fine-grained budget control from 32 to 256 tokens.

\textbf{Matched-Budget Reconstruction.}
For an apples-to-apples comparison at fixed token budgets, we evaluate AdaTok in non-adaptive mode at exactly $k\!\in\!\{32,64,128,256\}$ tokens and report rFID, PSNR, SSIM, and LPIPS (Tab.~\ref{tab:budget_sweep}). At 128 tokens AdaTok already reaches rFID \textbf{1.63} and LPIPS \textbf{0.216}, surpassing fixed-length TiTok-S at the same budget (rFID 1.71); at 256 tokens all four metrics improve further. The monotone trend across budgets confirms that prefix tokens carry an information hierarchy rather than redundant filler, and that AdaTok's gains hold under matched-budget protocols rather than relying on unequal token counts.

\textbf{Coarse-to-Fine Hierarchy Verification.}
To directly test the coarse-to-fine ordering NTM is meant to induce (Sec.~\ref{sec:prl}), we apply FFT spectral decomposition to 1K ImageNet validation reconstructions at each fixed budget and track the low-frequency energy ratio. It falls monotonically as the budget grows: $0.3826$ at 32 tokens, $0.3754$ at 64, $0.3710$ at 128, and $0.3704$ at 256, converging toward the source images' ratio of $0.3667$. Prefix tokens therefore carry low-frequency structural content while later tokens add progressively higher-frequency detail, supporting the information hierarchy PRL was designed to induce.

\begin{table}[t]
\centering
\small
\caption{\textbf{Ablation and failure-mode diagnosis.} NM: Nested Tail Masking; MH: MH-LoRA. Cases (b,d) expose naive representation and allocation failure modes; Case (e) is GRPO without Dynamic Pareto Weighting; Case (f) is full AdaTok. rFID on ImageNet-1K.}
\label{tab:ablation}
\setlength{\tabcolsep}{7pt}
\begin{tabular}{l cc c cc}
\hline
\textbf{Case} & \textbf{NM} & \textbf{MH} & \textbf{Policy} & \textbf{rFID}\,$\downarrow$ & \textbf{Tokens} \\ \hline
\multicolumn{6}{c}{\textit{PRL module study}} \\ \hdashline
(a) & \ding{55} & \ding{55} & Fixed & 1.71 & 128 \\
(b) & \ding{51} & \ding{55} & Fixed & 2.04 & 128 \\
(c) & \ding{51} & \ding{51} & Fixed & 1.63 & 128 \\
\hline
\multicolumn{6}{c}{\textit{policy study}} \\ \hdashline
(d) & \checkmark & \checkmark & Heur. & 2.32 & 155.3 \\
(e) & \checkmark & \checkmark & GRPO & 1.58 & 123 \\
\cellcolor{gray!20}(f) & \cellcolor{gray!20}\checkmark & \cellcolor{gray!20}\checkmark & \cellcolor{gray!20}\textbf{GRPO+DPW} & \cellcolor{gray!20}\textbf{1.50} & \cellcolor{gray!20}\textbf{118} \\ \hline
\end{tabular}
\vspace{-10pt}
\end{table}

\subsection{Ablation Study}

\textbf{PRL components (Cases a--c).} Adding \textit{Nested Tail Masking} (NM) on its own hurts the vanilla 1D baseline (Case a, equivalent to TiTok-S~\citep{yu2025image} at rFID 1.71 $\rightarrow$ Case b, 2.04), directly diagnosing semantic shift: one static decoder cannot use the same prefix both as a standalone structural code and as a scaffold for longer decodes. Adding \textit{MH-LoRA} (Case c) decouples these roles and recovers rFID \textbf{1.63}, surpassing the baseline. MH-LoRA is therefore the mechanism that makes ordered discrete prefixes usable across budgets.

\textbf{ATA vs.\ heuristic allocation (Cases d--f).} On the same PRL backbone, the heuristic policy (Case d) binary-searches $\mathcal{L}$ per image for the smallest budget under a fixed rMSE threshold, requiring multiple decoder passes and landing at 2.32 rFID with 155.3 tokens. GRPO without DPW improves this to 1.58 rFID at 123 tokens (Case e), while GRPO+DPW reaches \textbf{1.50} rFID at \textbf{118} tokens in one pass (Case f). The full allocator therefore improves both quality and length over the fixed 128-token counterpart (Case c, 1.63), which is the core adaptive-tokenization claim.

\textbf{Policy learning paradigms.} We further compare GRPO against REINFORCE~\citep{williams1992simple} and PPO~\citep{schulman2017proximal} on the same PRL backbone (Tab.~\ref{tab:policy_compare}). REINFORCE collapses to too-short sequences under high gradient variance, while PPO over-allocates and saturates the maximum budget. Only GRPO attains both the highest hypervolume in this comparison (HV=1342) and a healthy policy entropy ($H(\pi){=}0.28$), supporting the group-relative advantage as a suitable inductive bias for adaptive-budget tokenization.

\subsection{Further Analysis}

\begin{table}[t]
\centering
\small
\caption{\textbf{Matched-Budget Reconstruction of AdaTok.} AdaTok evaluated in non-adaptive mode at fixed budgets $k\!\in\!\{32,64,128,256\}$ on ImageNet-1K ($256{\times}256$). All four metrics improve monotonically with the budget. Adaptive mode (avg.\ $\sim$118 tokens) attains rFID \textbf{1.50}, beating the fixed-128 decoder (1.63) while still using fewer tokens on average; TiTok-S@128: 1.71 for reference.}
\label{tab:budget_sweep}
\setlength{\tabcolsep}{8pt}
\begin{tabular}{c cccc}
\toprule
\textbf{Tokens} & \textbf{rFID}\,$\downarrow$ & \textbf{PSNR}\,$\uparrow$ & \textbf{SSIM}\,$\uparrow$ & \textbf{LPIPS}\,$\downarrow$ \\
\midrule
32  & 4.60 & 16.48 & 0.366 & 0.330 \\
64  & 2.64 & 17.27 & 0.395 & 0.262 \\
128 & 1.63 & 17.92 & 0.421 & 0.216 \\
256 & \textbf{1.31} & \textbf{18.42} & \textbf{0.434} & \textbf{0.196} \\
\bottomrule
\end{tabular}
\vspace{-8pt}
\end{table}

\begin{table}[t]
\centering
\begin{minipage}[t]{0.50\linewidth}
\centering
\small
\captionof{table}{\textbf{Policy Learning Paradigms.} HV: Hypervolume; $H(\pi)$: policy entropy. GRPO attains the highest hypervolume in this comparison without an auxiliary critic.}
\label{tab:policy_compare}
\setlength{\tabcolsep}{3pt}
\begin{tabular}{l cc c cc}
\toprule
& \multicolumn{2}{c}{\textbf{Policy}} & & \multicolumn{2}{c}{\textbf{Recon.}} \\
\cmidrule(lr){2-3} \cmidrule(lr){5-6}
\textbf{Strategy} & \textbf{HV}\,$\uparrow$ & \textbf{$H(\pi)$}\,$\uparrow$ & & \textbf{rMSE}\,$\downarrow$ & \textbf{Tok.} \\
\midrule
Fixed (128) & 226 & 0.00 & & 0.0219 & 128.0 \\
REINFORCE & 182 & 0.08 & & 0.0263 & \textbf{36.8} \\
PPO & 419 & 0.0 & & \textbf{0.0197} & 224 \\
\rowcolor{gray!20}
\textbf{GRPO} & \textbf{1342} & \textbf{0.28} & & 0.0232 & 118 \\
\bottomrule
\end{tabular}
\end{minipage}\hfill
\begin{minipage}[t]{0.46\linewidth}
\centering
\small
\captionof{table}{\textbf{AR Generation, Matched Compute.} 195M LlamaGen-style AR, 200K steps. Speedup is theoretical length reduction $256/k$; $\Delta$gFID is relative to the 256-token decode (self-comparison only).}
\label{tab:ar_pareto}
\setlength{\tabcolsep}{4pt}
\begin{tabular}{c c c c}
\toprule
\textbf{Tok.} & \textbf{Speedup}\,$\uparrow$ & \boldmath{$\Delta$\textbf{gFID}}\,$\downarrow$ & \textbf{gFID} \\
\midrule
32  & $8\times$ & $+2.60$ & 15.07 \\
64  & $4\times$ & $+0.97$ & 13.44 \\
128 & $2\times$ & $\mathbf{-0.34}$ & 12.13 \\
256 & $1\times$ & $0$ (ref.) & 12.47 \\
\bottomrule
\end{tabular}
\end{minipage}
\vspace{-8pt}
\end{table}

\textbf{Optimization trajectory.} Fig.~\ref{fig:trajectory} traces the ATA learning trajectory: GRPO with dynamic Pareto weighting reaches a better rate--distortion trade-off than fixed-weight scalarization, confirming that input-dependent reweighting helps over a static $\lambda$.

\textbf{AR generation under matched compute.}\label{sec:ar_pareto}
Separately from the MaskGIT compatibility result in Tab.~\ref{tab:rec}, we train a single 195M LlamaGen-style AR generator on AdaTok tokens for 200K steps and vary the inference-time budget; this exposes a clean throughput--quality trade-off (Tab.~\ref{tab:ar_pareto}). Because the matched-compute setup deliberately undertrains the generator, absolute gFID is not comparable to fully-trained AR baselines, so we report \emph{relative} $\Delta$gFID against the 256-token decode. In this self-comparison, 128 tokens already \emph{improve} on the 256-token decode ($\Delta\!=\!-0.34$) while halving inference cost, and 32 tokens trade a modest quality drop ($\Delta\!=\!+2.60$) for an $8{\times}$ length reduction. The measured $\sim$$2.1{\times}$ wall-clock throughput at the 118-token decode reports the speedup of decoding to a fixed 118-token target chosen to match AdaTok-Adaptive's average; generation-time autonomous selection without access to the target image requires a visual end-of-sequence mechanism that we leave to future work.
\textbf{Computational efficiency.}
\label{sec:efficiency}
We measure costs against a fixed 256-token baseline on a single H100. Training overhead is small: GRPO trains a 330K-parameter MLP on top of a frozen tokenizer in 10K steps ($\sim$3\% of total) and needs no value network, while MH-LoRA adds only $\sim$1.1\% parameters. Tokenizer-stage speedup is modest ($\sim$12--17\% across batch sizes, limited by 31\% padding); the dominant gain shows up at AR generation, where shorter sequences translate linearly into throughput: $\sim$$2.1{\times}$ at the 118-token average and $8{\times}$ at 32 tokens (Sec.~\ref{sec:ar_pareto}).

\textbf{On the novelty of each component.}
Each module uses familiar ingredients, but the learned object is new: a discrete visual tokenizer whose token count is an input-conditioned output. This requires all three pieces to close the loop. Ordered prefixes make multiple budgets meaningful; MH-LoRA removes the Semantic Shift that otherwise breaks those prefixes (Tab.~\ref{tab:ablation}, b vs.\ c); deterministic-group GRPO+DPW then chooses the prefix in one pass and improves both quality and length over fixed or GRPO-only allocation (f vs.\ c/e). The novelty is therefore not variable-length reconstruction alone, but making elasticity operational inside a discrete tokenizer without search, external predictors, or target-quality conditioning.

\section{Conclusion}
\label{sec:conclusion}
We presented AdaTok, a self-budgeting discrete 1D tokenizer that learns token count as an input-conditioned output. PRL resolves Semantic Shift with budget-specific MH-LoRA heads, while deterministic-group GRPO+DPW turns the candidate budget set into an exact relative-reward group for one-pass allocation. On ImageNet-1K, AdaTok-Adaptive reaches rFID 1.50 at $\sim$118 average tokens, AdaTok-Full attains gFID 2.28 at 256 tokens, and AR throughput at the 118-token average is $\sim$$2.1{\times}$ over a fixed 256-token baseline. The codebook plugs directly into MaskGIT and LlamaGen-style generators, and we plan to release code and checkpoints.

\newpage
\bibliographystyle{plainnat}
\bibliography{reference}

@article{dhariwal2021diffusion,
  title={Diffusion models beat gans on image synthesis},
  author={Dhariwal, Prafulla and Nichol, Alexander},
  journal={Advances in neural information processing systems},
  volume={34},
  pages={8780--8794},
  year={2021}
}

@inproceedings{rombach2022high,
  title={High-resolution image synthesis with latent diffusion models},
  author={Rombach, Robin and Blattmann, Andreas and Lorenz, Dominik and Esser, Patrick and Ommer, Bj{\"o}rn},
  booktitle={Proceedings of the IEEE/CVF conference on computer vision and pattern recognition},
  pages={10684--10695},
  year={2022}
}

@inproceedings{esser2021taming,
  title={Taming transformers for high-resolution image synthesis},
  author={Esser, Patrick and Rombach, Robin and Ommer, Bjorn},
  booktitle={Proceedings of the IEEE/CVF conference on computer vision and pattern recognition},
  pages={12873--12883},
  year={2021}
}

@article{sun2024autoregressive,
  title={Autoregressive model beats diffusion: Llama for scalable image generation},
  author={Sun, Peize and Jiang, Yi and Chen, Shoufa and Zhang, Shilong and Peng, Bingyue and Luo, Ping and Yuan, Zehuan},
  journal={arXiv preprint arXiv:2406.06525},
  year={2024}
}

@inproceedings{kingma2013auto,
  title={Auto-Encoding Variational Bayes},
  author={Kingma, Diederik P. and Welling, Max},
  booktitle={2nd International Conference on Learning Representations (ICLR)},
  year={2014},
  url={https://arxiv.org/abs/1312.6114},
  eprint={1312.6114},
  archivePrefix={arXiv}
}

@article{tian2024visual,
  title={Visual autoregressive modeling: Scalable image generation via next-scale prediction},
  author={Tian, Keyu and Jiang, Yi and Yuan, Zehuan and Peng, Bingyue and Wang, Liwei},
  journal={Advances in neural information processing systems},
  volume={37},
  pages={84839--84865},
  year={2024}
}

@article{van2017neural,
  title={Neural discrete representation learning},
  author={Van Den Oord, Aaron and Vinyals, Oriol and others},
  journal={Advances in neural information processing systems},
  volume={30},
  year={2017}
}

@article{yu2025image,
  title={An image is worth 32 tokens for reconstruction and generation},
  author={Yu, Qihang and Weber, Mark and Deng, Xueqing and Shen, Xiaohui and Cremers, Daniel and Chen, Liang-Chieh},
  journal={Advances in Neural Information Processing Systems},
  volume={37},
  pages={128940--128966},
  year={2025}
}

@inproceedings{duggal2024adaptive,
  title={Adaptive length image tokenization via recurrent allocation},
  author={Duggal, Shivam and Isola, Phillip and Torralba, Antonio and Freeman, William T},
  booktitle={First Workshop on Scalable Optimization for Efficient and Adaptive Foundation Models},
  year={2024}
}

@article{duggal2025karl,
  author       = {Shivam Duggal and
                  Sanghyun Byun and
                  William T. Freeman and
                  Antonio Torralba and
                  Phillip Isola},
  title        = {Single-pass Adaptive Image Tokenization for Minimum Program Search},
  journal      = {CoRR},
  volume       = {abs/2507.07995},
  year         = {2025},
  url          = {https://doi.org/10.48550/arXiv.2507.07995},
  doi          = {10.48550/ARXIV.2507.07995}
}

@article{shao2024deepseekmath,
  author       = {Zhihong Shao and
                  Peiyi Wang and
                  Qihao Zhu and
                  Runxin Xu and
                  Junxiao Song and
                  Mingchuan Zhang and
                  Y. K. Li and
                  Y. Wu and
                  Daya Guo},
  title        = {{DeepSeekMath}: Pushing the Limits of Mathematical Reasoning in Open
                  Language Models},
  journal      = {CoRR},
  volume       = {abs/2402.03300},
  year         = {2024},
  url          = {https://doi.org/10.48550/arXiv.2402.03300},
  doi          = {10.48550/ARXIV.2402.03300}
}

@article{williams1992simple,
  title={Simple statistical gradient-following algorithms for connectionist reinforcement learning},
  author={Williams, Ronald J},
  journal={Machine learning},
  volume={8},
  pages={229--256},
  year={1992},
  publisher={Springer}
}

@article{schulman2017proximal,
  title={Proximal policy optimization algorithms},
  author={Schulman, John and Wolski, Filip and Dhariwal, Prafulla and Radford, Alec and Klimov, Oleg},
  journal={arXiv preprint arXiv:1707.06347},
  year={2017}
}

@article{lu2025learning,
  title={Learning to optimize multi-objective alignment through dynamic reward weighting},
  author={Lu, Yining and Wang, Zilong and Li, Shiyang and Liu, Xin and Yu, Changlong and Yin, Qingyu and Shi, Zhan and Zhang, Zixuan and Jiang, Meng},
  journal={arXiv preprint arXiv:2509.11452},
  year={2025}
}

@inproceedings{Maskgit,
  title={Maskgit: Masked generative image transformer},
  author={Chang, Huiwen and Zhang, Han and Jiang, Lu and Liu, Ce and Freeman, William T},
  booktitle={Proceedings of the IEEE/CVF conference on computer vision and pattern recognition},
  pages={11315--11325},
  year={2022}
}

@article{miwa2025one,
  title={One-D-Piece: Image Tokenizer Meets Quality-Controllable Compression},
  author={Miwa, Keita and Sasaki, Kento and Arai, Hidehisa and Takahashi, Tsubasa and Yamaguchi, Yu},
  journal={arXiv preprint arXiv:2501.10064},
  year={2025}
}

@article{shen2025cat,
  title={Cat: Content-adaptive image tokenization},
  author={Shen, Junhong and Tirumala, Kushal and Yasunaga, Michihiro and Misra, Ishan and Zettlemoyer, Luke and Yu, Lili and Zhou, Chunting},
  journal={arXiv preprint arXiv:2501.03120},
  year={2025}
}

@inproceedings{wen2025principal,
  title={" Principal Components" Enable A New Language of Images},
  author={Wen, Xin and Zhao, Bingchen and Elezi, Ismail and Deng, Jiankang and Qi, Xiaojuan},
  booktitle={Proceedings of the IEEE/CVF International Conference on Computer Vision},
  pages={16641--16651},
  year={2025}
}

@article{bachmann2025flextok,
  title={FlexTok: Resampling Images into 1D Token Sequences of Flexible Length},
  author={Bachmann, Roman and Allardice, Jesse and Mizrahi, David and Fini, Enrico and Kar, O{\u{g}}uzhan Fatih and Amirloo, Elmira and El-Nouby, Alaaeldin and Zamir, Amir and Dehghan, Afshin},
  journal={arXiv preprint arXiv:2502.13967},
  year={2025}
}

@article{yan2024elastictok,
  title={Elastictok: Adaptive tokenization for image and video},
  author={Yan, Wilson and Mnih, Volodymyr and Faust, Aleksandra and Zaharia, Matei and Abbeel, Pieter and Liu, Hao},
  journal={arXiv preprint arXiv:2410.08368},
  year={2024}
}

@article{yang2022visual,
  title={Visual concepts tokenization},
  author={Yang, Tao and Wang, Yuwang and Lu, Yan and Zheng, Nanning},
  journal={Advances in Neural Information Processing Systems},
  volume={35},
  pages={31571--31582},
  year={2022}
}

@inproceedings{rippel2014learning,
  title={Learning ordered representations with nested dropout},
  author={Rippel, Oren and Gelbart, Michael and Adams, Ryan},
  booktitle={International Conference on Machine Learning},
  pages={1746--1754},
  year={2014},
  organization={PMLR}
}

@article{rao2021dynamicvit,
  title={Dynamicvit: Efficient vision transformers with dynamic token sparsification},
  author={Rao, Yongming and Zhao, Wenliang and Liu, Benlin and Lu, Jiwen and Zhou, Jie and Hsieh, Cho-Jui},
  journal={Advances in neural information processing systems},
  volume={34},
  pages={13937--13949},
  year={2021}
}

@inproceedings{yin2022vit,
  title={A-vit: Adaptive tokens for efficient vision transformer},
  author={Yin, Hongxu and Vahdat, Arash and Alvarez, Jose M and Mallya, Arun and Kautz, Jan and Molchanov, Pavlo},
  booktitle={Proceedings of the IEEE/CVF conference on computer vision and pattern recognition},
  pages={10809--10818},
  year={2022}
}

\end{document}